\documentclass[letterpaper, 10 pt, journal, twoside]{IEEEtran}

\usepackage{amsmath,amsfonts}
\usepackage{algorithmic}
\usepackage{algorithm}
\usepackage{array}
\usepackage[caption=false,font=normalsize,labelfont=sf,textfont=sf]{subfig}
\usepackage{textcomp}
\usepackage{url}
\usepackage{verbatim}
\usepackage{graphicx}
\usepackage{cite}
\usepackage{lipsum}
\usepackage{xcolor}
\usepackage{soul}
\usepackage{doi}
\usepackage{caption}
\usepackage{multirow}
\usepackage{grffile}
\usepackage{stfloats}
\usepackage{booktabs}

\newcommand{\sd}[1]{{\,\tiny #1}}            
\newcommand{\rx}[1]{\textcolor{gray}{#1}}    

\usepackage{hyperref} 
\hypersetup{
    colorlinks,
    citecolor=black,
    filecolor=black,
    linkcolor=black,
    urlcolor=black
}
\usepackage[all]{hypcap}

\usepackage{notoccite}
\usepackage{etoolbox}
\makeatletter
\newlength{\topfloatoffset}
\newcommand{\topfloatpatchfailed}{\PackageError{main}{Top float offset patch failed}{}}
\advance\textfloatsep\topfloatoffset
\patchcmd{\@cflt}{\setbox\@tempboxa \vbox{}}{\setbox\@tempboxa \vbox{\vskip\topfloatoffset}}{}{\topfloatpatchfailed}
\patchcmd{\@cflt}{\vskip \textfloatsep}{\vskip \glueexpr\textfloatsep-\topfloatoffset\relax}{}{\topfloatpatchfailed}
\newcommand{\dbltopfloatoffset}{\ifnum\c@page=\@ne\z@\else\topfloatoffset\fi}
\newcommand{\enabledbltopfloatoffset}{\global\advance\dbltextfloatsep\topfloatoffset}
\patchcmd{\@cdblflt}{\setbox\@tempboxa \vbox{}}{\setbox\@tempboxa \vbox{\vskip\dbltopfloatoffset}}{}{\topfloatpatchfailed}
\patchcmd{\@cdblflt}{\vskip \dbltextfloatsep}{\vskip \glueexpr\dbltextfloatsep-\topfloatoffset\relax}{}{\topfloatpatchfailed}
\patchcmd{\@cflb}{\vskip \textfloatsep}{\vskip \glueexpr\textfloatsep-\topfloatoffset\relax}{}{\topfloatpatchfailed}
\patchcmd{\@cflb}{\vskip -\floatsep}{\vskip -\floatsep \vskip\topfloatoffset}{}{\topfloatpatchfailed}
\patchcmd{\@cdblflb}{\vskip \dbltextfloatsep}{\vskip \glueexpr\dbltextfloatsep-\topfloatoffset\relax}{}{\topfloatpatchfailed}
\patchcmd{\@cdblflb}{\vskip -\dblfloatsep}{\vskip -\dblfloatsep \vskip\topfloatoffset}{}{\topfloatpatchfailed}
\makeatother

\begin{document}
\bstctlcite{IEEEexample:BSTcontrol}

\title{ARENA: Adaptive Risk-aware Navigation for Autonomous Aerial Inspection}

\author{David-Alexandre Poissant$^{1}$, Alexis Lussier Desbiens$^{2}$, François Ferland$^{3}$, and Louis Petit$^{1}$
\thanks{This research was funded by NSERC and CRIAQ in partnership with Hydro-Québec and DRONE VOLT®. The authors are with Université de Sherbrooke (UdeS).}
\thanks{$^{1}$First and last authors are with the Sherbrooke Autonomous Field Robotics Lab (SAFiR Lab), UdeS, {\tt\small david-alexandre.poissant@usherbrooke.ca}}\thanks{$^{2}$Second author is with the Createk Design Lab, UdeS.}
\thanks{$^{3}$Third author is with the Intelligent Interactive Integrated Interdisciplinary Robot Lab (IntRoLab), UdeS.}
}



\maketitle
\enabledbltopfloatoffset 

\begin{abstract}
Autonomous robotic outdoor inspection missions require balancing multiple conflicting objectives while navigating near dangerous obstacles. Current multi-objective path planning (MOPP) methods struggle to adapt to evolving risks like localization errors, weather, battery state, and communication issues. This letter presents the Adaptive Risk-aware Navigation (ARENA) MOPP approach for UAVs in complex 3D environments. This method enables online trajectory adaptation by optimizing safety, time, and energy consumption using a 4D NURBS representation and a genetic-based algorithm to generate the Pareto front, a novel semi-empirical energy consumption model for various flight situations, and a novel risk-aware voting algorithm ensuring adaptivity. Tests in multiple simulated environments and one real-world scenario demonstrate the planner's ability to produce diverse, optimized trajectories covering 88.4\% $\pm$ 5.5\% of the range defined by high-fidelity benchmarks and a mean estimation error of 14\% for power consumption. The ARENA framework enhances UAV autonomy and reliability in critical, evolving 3D missions. More details can be found at \url{https://safirlab.github.io/research/arena/}.
\end{abstract}


\section{Introduction}
\IEEEPARstart{U}{ncrewed} aerial vehicles (UAVs) are becoming crucial tools in various scenarios where human involvement can become too risky or incur high costs, such as search and rescue \cite{khan2021}, surveillance \cite{shakhatreh2019}, and inspection \cite{leclerc2023, petit2022}. Achieving autonomy in these scenarios heavily relies on the path planning module to generate safe and feasible trajectories. Numerous approaches have been proposed to find the shortest or safest path in a cluttered environment. For intricate tasks, such as infrastructure inspection, ensuring safe maneuvers involves considering multiple factors, prompting increased interest in multi-objective optimization (MOO) methods.

\begin{figure}[!t]
    \centering
    \captionsetup[subfloat]{textfont=small, width=0.48\linewidth}
    \subfloat[High winds might cause deviations, risking arcing or collision]{%
        \includegraphics[width=0.48\linewidth]{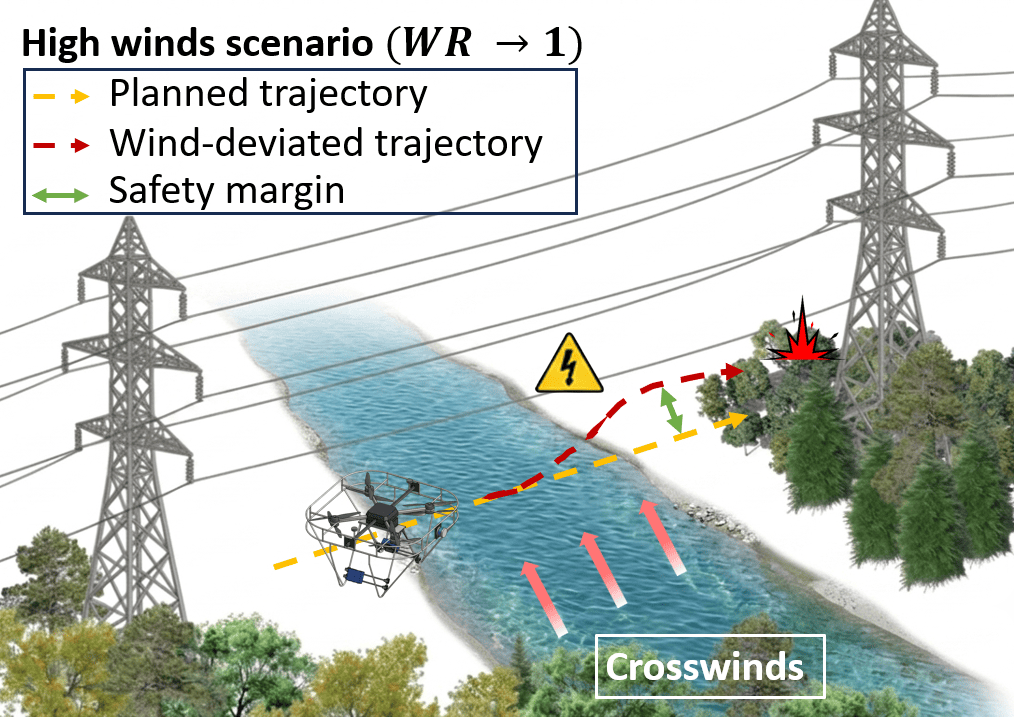}
        \label{fig:high_winds_scenario}%
    }
    \subfloat[Low localization precision might reduce position precision, risking arcing or collision]{%
        \includegraphics[width=0.48\linewidth]{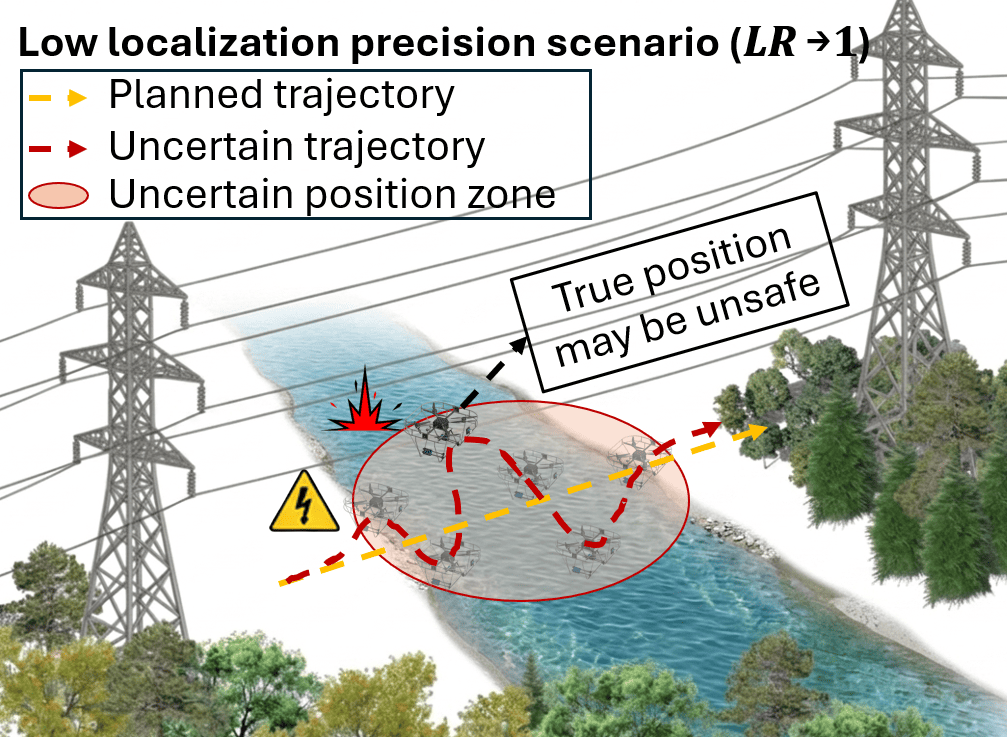}
        \label{fig:low_localization_precision_scenario}%
    } \\
    \subfloat[Low battery might occur during mission, risking crash or collision]{%
        \includegraphics[width=0.48\linewidth]{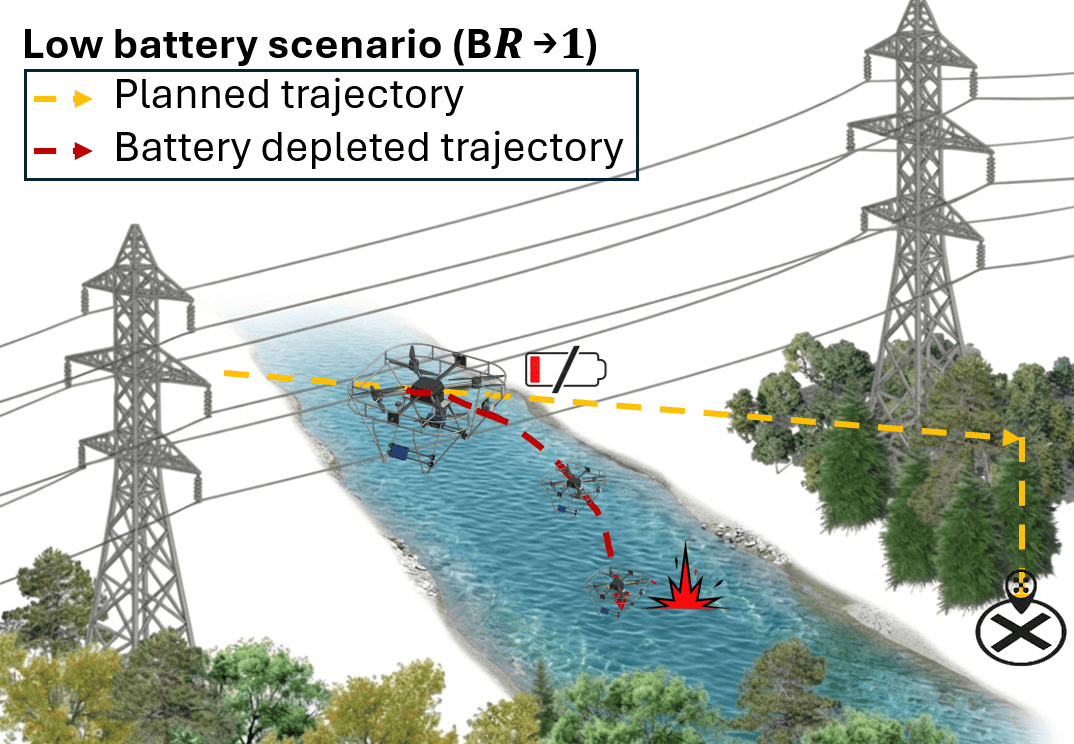}
        \label{fig:low_battery_scenario}%
    }
    \subfloat[Low communication can reduce trust in autonomous mission execution]{%
        \includegraphics[width=0.48\linewidth]{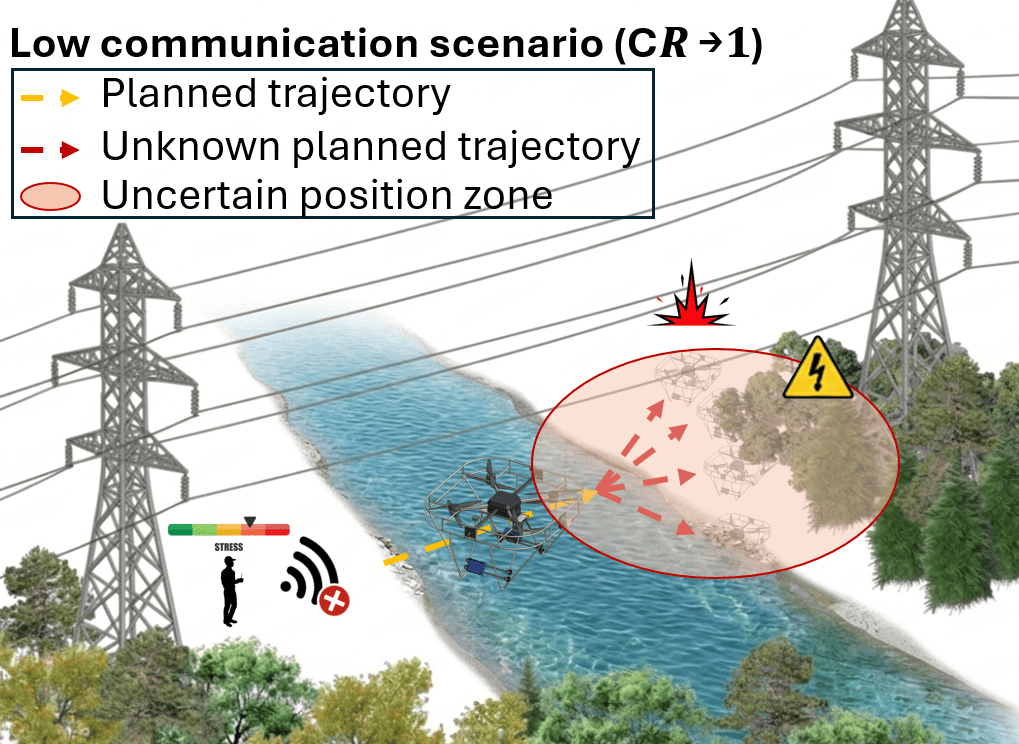}
        \label{fig:low_communication_scenario}%
    }
    \captionsetup{width=\linewidth}
    \caption{High risk scenarios for a power line inspection drone}
    \label{fig:risk_scenarios}
    \vspace*{-\baselineskip}
\end{figure}


When path planning involves multiple objectives, there is generally no single optimal solution but rather a set of trade-off paths forming the Pareto front, which MOO aims to identify. To the authors' knowledge, most Multi-Objective Path Planning (MOPP) methods include static costs like safety, time, smoothness, or others to find the optimal front \cite{hayes2022}. Some research estimates risks using offline data, such as satellite imagery and building positions \cite{causa2023}, but lacks dynamic adaptation to mission conditions. The recent Multi-Objective and Adaptive Risk-aware (MOAR) path planner \cite{petit2024} enables real-time trajectory modulation by considering evolving risks, like the scenarios shown in Fig. \ref{fig:risk_scenarios}. However, it relies on exhaustive graph searches in discretized 2D environments, limiting real-time applicability in large 3D spaces. This letter proposes an extended version of the MOAR path planner, broadening its use to 3D spaces and continuous representations. The proposed framework, named Adaptive Risk-aware Navigation (ARENA), leverages a 4D Non-uniform rational B-spline (NURBS) \cite{piegl1995} representation and the non-dominated sorting genetic algorithm (NSGA-II) \cite{deb2002} to compute feasible trajectories and modulate the velocity profile along them. It optimizes safety, time, and a novel semi-empirical energy objective while adapting to evolving risks like localization precision, wind, battery state, and communication. The framework's reliability and efficiency are demonstrated through numerous simulations and real-world scenarios. Although developed primarily for power line inspection, we further validated it in a randomly cluttered environment, showing that the algorithm can be applied to a broad range of drone path planning problems.

Section \ref{sec:related_work} reviews the state of the art on MOPP. Section \ref{sec:fundamentals} explains the fundamentals of MOO and NURBS to provide a clearer understanding of the approach described in Section \ref{sec:risk_aware_path_planning}. Section \ref{sec:results} shows the results of the sensitivity analysis and risk adaptability. Section \ref{sec:conclusion} concludes the letter by suggesting potential future work.

\section{Related Work}
\label{sec:related_work}

\paragraph{Multi-Objective Optimization and Risk Adaptivity} MOPP methods are widely used for global trajectory generation in UAV applications, addressing various objectives \cite{hayes2022}. A common approach to handling multiple objectives combines cost functions into a single scalar value. This typically involves weighing the objectives to obtain desirable behavior. Previous work applied this method by integrating safety, energy, and time \cite{petit2024}. Similarly, the authors in \cite{weiss2006} combined path length with a risk-utility function based on a risk map. Zhou et al. \cite{zhou2021} introduced a gradient-based method that incorporates smoothness and feasibility functions to local graph search-based collision risks. While this approach simplifies the inherently multi-objective problems, the authors in \cite{hayes2022} highlight several issues with this workflow. One key concern is that preferences between objectives may evolve over time and missions, making the transformation of an inherently multi-objective problem into a single cost function a semi-blind, manual process. The authors propose another approach to handling multiple objectives in \cite{korpan2023}, where they use A* independently on every objective to mimic specific behaviors and then use a voting algorithm to choose between them. While this method produces paths at the extreme edge of the Pareto front, it fails to address the key issue of evolving preferences between objectives over time, and does not allow for compromises that could better align with a specific situation.

\paragraph{Risk Assessment} Conducting a thorough risk assessment is essential to establish trust in UAV autonomous operations. Some studies introduce safety objectives to prevent unsafe trajectories during MOO processes. Ahmed et al. \cite{ahmed2011, ahmed2013} assess trajectory safety using the robot's range of vision. Many approaches maximize safety by minimizing collision risks with static or dynamic obstacles using clearance-based objectives \cite{khan2021}. Authors in \cite{weiss2006} use collision risk maps, while \cite{rubiohervas2018} combines Gaussian processes and Evolutionary algorithms (EA) to map risks like signal strength and wind speed for single or multiple UAVs. However, these methods address fixed mission statements. Preferences between objectives can shift dynamically during and between missions. Factors such as worsening weather or GPS reception may prioritize safety, while battery constraints could favor faster or more energy-efficient paths. Recent work introduced the concepts of damage cost in the event of a crash and non-insertion cost, which quantifies the degree of intrusion into important obstacle structures \cite{petit2024}, providing situational awareness and a margin of error for effective response to hazardous situations.

\paragraph{Trajectory Representation} While not tied to a specific mission statement, the two concepts of damage and non-insertion costs in \cite{petit2024} are constrained to a 2D plane with discretely constructed paths, similar to \cite{song2019, primatesta2019}, leading to resolution loss and limited path optimality. To overcome this, \cite{zhou2021} proposes B-spline parametrization to represent trajectories continuously, using knot vectors based on actuator limitations. However, this method lacks flexibility in modulating trajectories. For instance, control points are positioned at fixed intervals, constraining the optimizer's ability to discover more optimal solutions. Another method \cite{cheatham2005} proposes an n-dimensional NURBS approach that adds orientation components to XYZ control points. While this enables smoothing additional components of a Euclidean path, it relies on a look-ahead process to ensure compliance with actuator limitations. However, post-optimization adjustments undermine the optimization purpose, as they may deviate from the originally optimized solution.

\paragraph{Energy Efficiency} Energy efficiency is crucial for UAV missions, especially under battery constraints. Several methods exist to model energy consumption \cite{zhang2021}. For example, \cite{zeng2017} defines a detailed energy model for fixed-wing UAVs based on speeds and accelerations along trajectories. Other methods, like \cite{hasini2018}, empirically derive energy consumption by evaluating a drone under various scenarios to create situational equations. While validated through theoretical energy consumption, this approach depends on best-fitting curves rather than physics-based principles, making it scenario-dependent. To the authors' knowledge, no method formulates a generalized data-driven energy model using physics-based principles.

This paper introduces a novel framework for an online adaptive energy-efficient 3D multi-objective path planning algorithm designed to generate an approximate Pareto set of trajectories that accommodate risk variation during a mission. The contributions are as follows: (a) a novel voting algorithm for choosing the best trajectory according to evolving mission risks, (b) an extension of the notion of non-insertion cost introduced in recent work \cite{petit2024}, (c) the introduction of a 4$^{th}$ dimension in a continuous representation tool to modulate and optimize speed while complying with actuator constraints, and (d) a novel semi-empirical energy consumption model that incorporates different flight situations.
\section{Fundamentals}
\label{sec:fundamentals}

This section introduces the concepts required to understand the algorithm in Section \ref{sec:risk_aware_path_planning}: the definition of a MOO process and the trajectory representation tool used in ARENA.

\subsection{Multi-objective Optimization}

The goal in MOO is to find a \textit{D}-dimensional solution vector
$z = [z_{1}, \dots, z_{D}]^{T}$, 
that minimizes or maximizes a set $\mathcal{F} = \{f_{e}(z) \mid e \in \{1, \dots, E\}\}$ of $E$ objective functions, that can be subject to $F$ inequality constraints $g_{f}(z) \geq 0, \ \forall f \in \{1, \dots, F\}$,
as well to $G$ equality constraints $h_{g}(z) = 0, \ \forall g \in \{1, \dots, G\}$.
Additionally, each element of $z$ can be limited by a lower and upper bound $z_{d}^{(L)} \leq z_{d} \leq z_{d}^{(U)}, \ \forall d \in \{1, \dots, D\}$.

The search space $S$ contains all feasible solutions to the optimization problem. A solution is possible if it satisfies all constraints and bounds. Feasible solutions can be ranked by dominance, as no single solution may optimize all objectives simultaneously. A solution $z_{1}$ is said to dominate another solution $z_{2}$ $(z_{1} \preceq z_{2})$ if $z_{1}$ is equal or better than $z_{2}$ for all objectives and strictly better in at least one. Each non-dominated solution forms the Pareto set.

\subsection{Non-Uniform Rational B-Splines (NURBS)}
\label{sec:nurbs}

A NURBS curve of order $p + 1$ is defined by Piegl and Tiller \cite{piegl1995} as
\begin{equation}
    \label{nurbs_equation}
    C(u) = \frac{\sum_{i=0}^{n} N_{i, p}(u)w_{i}P_{i}}{\sum_{i=0}^{n} N_{i, p}(u)w_{i}}
\end{equation}
where $n + 1$ is the number of control points, $p$ is the degree of the basis function $N_{i, p}$, $P_{i} = [x_{i} \ \ y_{i} \ \ z_{i}]^{T}$ is the $i^{th}$ control point (assuming a 3D curve), and $w_{i}$ is the associated weight. The basis functions $N_{i, p}$ are defined with respect to the parameter $u$ and a fixed knot vector
$U = [u_{0} \ \ ... \ \ u_{m}]^{T}$, containing $m + 1$ knots, where $m = n + p + 1$. The Cox-de Boor formula computes the basis functions recursively \cite{piegl1995}. 
\section{Adaptive Risk-Aware Path Planning}
\label{sec:risk_aware_path_planning}

\begin{figure*}
    \centering
    \includegraphics[width=\linewidth]{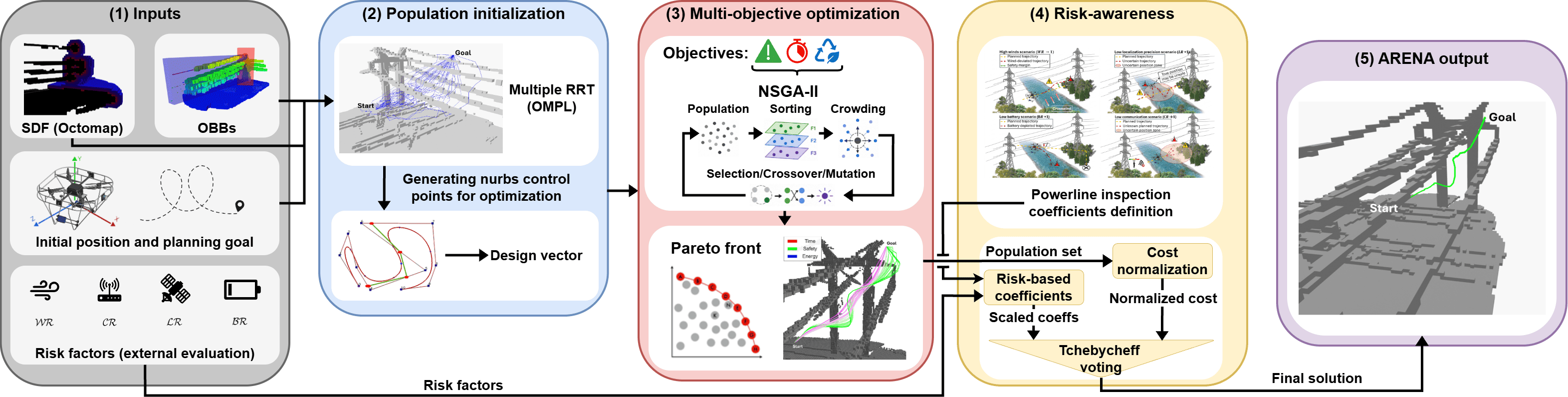}
    \caption{Overview of the proposed ARENA planning framework for risk-aware UAV navigation}
    \label{fig:arena_pipeline}
    \vspace*{-\baselineskip}
\end{figure*}

This section begins by defining the multi-objective path planning problem. The decision vector, continuous path representation method, and objectives are presented in Section \ref{sec:risk_aware_path_planning_problem_definition}. Section \ref{sec:risk_aware_path_planning_solver} describes the adaptive risk-aware multi-objective path planning method, including the multi-objective solver and its initialization, the voting algorithm, the real-time adjustment of its coefficients, and the hyperparameters of the entire path planning method.

\subsection{Problem Definition and Representation}
\label{sec:risk_aware_path_planning_problem_definition}

The infrastructure inspection path planning scenario is generalized as a 4D problem, defined as $\mathbb{D}^{4} = \{ (x, y, z, v) \mid x \in [x_{min}, x_{max}], y \in [y_{min}, y_{max}], z \in [z_{min}, z_{max}], v \in [-v_{max}, v_{max}] \}$ with start and goal positions $\{(x_{a}, y_{a}, z_{a}, v_{a}), (x_{b}, y_{b}, z_{b}, v_{b})\} \in \mathbb{D}^{4}$. The goal of this MOPP problem is to find the NURBS curve $C = \{C(u): u \in [a, b]\}$ that minimizes the set of objectives $\mathcal{F}$ described below.

Trajectories are represented using NURBS \cite{piegl1995}, described in Section \ref{sec:nurbs} and shown in Fig. \ref{fig:arena_pipeline}, block (2), to benefit from their inherent properties for multidimensional multi-objective optimization. These include \textit{strong convex hull} properties, \textit{local approximation} capability, and \textit{infinite differentiability} apart from knot multiplicity \cite{piegl1995}. The first two properties are helpful during optimization to constrain the curve inside a bounded area and to escape local minima by varying the curve locally. The last property ensures the smoothness of the curve according to the knot multiplicity of the curve, which is beneficial for UAVs that have strict actuator limitations, such as low acceleration. The design vector is given by:
\begin{equation}
    \label{eq:design_vector}
    \begin{aligned}
        z ={} & [w_{0} \ \ x_{1} \ \ y_{1} \ \ z_{1} \ \left \| \vec{v}_{1} \right \| \ w_{1} \ \ ... \\
              &x_{n-1} \ \ y_{n-1} \ \ z_{n-1} \ \left \| \vec{v}_{n-1} \right \| \ w_{n-1} \ \ w_{n}]^{T}
    \end{aligned}
\end{equation}
The 4$^{th}$ dimension is added to the parametric curve $C$, representing the velocity norm at each control point, which enables velocity profile modulation along trajectories as risks evolve. The start and goal positions, along with their speeds, are fixed and excluded from the decision vector $z$. The problem is constrained by two hard limits: acceleration ($a_{max}$ in Table \ref{table:parameters_table}) to meet actuator limitations and collision avoidance to ensure feasibility.

Costs are generalized into three functions to encompass multiple scenarios: time, safety, and energy consumption, which are described in the following subsections.

\subsubsection{Time cost}
\label{sec:time_cost}
The time cost is computed as:
\begin{equation}
    \label{eq:time_cost}
    \mathcal{F}_{Time} = \sum_{i=0}^{Q-1} \frac{d_{i}}{ \left \| \vec{v}_{i + 1} \right \| }
\end{equation}
where $Q$ is the number of sample points, and $d_{i}$ is the distance on the path segment. The $i + 1$ velocity is used, assuming inspection drones are slow and can quickly reach low speeds. Under normal conditions, with no significant risks during an inspection mission, the algorithm minimizes this cost function, prioritizing speeding up the inspection process over safety or energy consumption.

\subsubsection{Safety cost}
\label{sec:safety_cost}
The safety cost consists of two terms:
\begin{equation}
    \label{eq:safety_cost}
    \begin{aligned}
        \mathcal{F}_{Safety}={} & k_{a} (\frac{\sum_{i=0}^{Q} \mathcal{F}_{sdf_{i}}}{Q} + \max(\mathcal{F}_{sdf})) \\ & + k_{b} (\frac{\sum_{i=0}^{Q} \mathcal{F}_{ch_{i}}}{Q} + \max(\mathcal{F}_{ch}))
    \end{aligned}
\end{equation}
where $\mathcal{F}_{sdf_{i}}$ is a collision cost to ensure a safe UAV-obstacle distance and $\mathcal{F}{ch_{i}}$ denotes a non-insertion cost for obstacles forming convex hulls to keep trajectories outside critical zones. In this letter, convex hulls, modeled as oriented bounding boxes (OBB), are dynamically adjusted around cables and pylons using semantic information during power line inspections. The coefficients $k_{a}$ and $k_{b}$ in Eq. \ref{eq:safety_cost} are normalized since $\mathcal{F}{sdf_{i}}$ and $\mathcal{F}{ch_{i}}$ return normalized values. Combining mean and maximum costs ensures overall trajectory safety while maximizing the minimum obstacle distance. To support this, a signed distance field (SDF) \cite{jones2006} maintains obstacle information within the free space. $\mathcal{F}_{sdf_{i}}$ is defined using the closest obstacle distance in the SDF:
\begin{equation}
    \label{eq:sdf_cost}
    \mathcal{F}_{sdf_{i}} = 
    \begin{cases} 
        0 & d_{obs_{i}} \geq r_{sdf_{max}} \\
        \lambda \left(\frac{1}{d_{obs_{i}}} - \frac{1}{r_{sdf_{max}}}\right) & \text{otherwise} \\
        1 & d_{obs_{i}} \leq r_{sdf_{min}}
    \end{cases}
\end{equation}
where $\lambda = \frac{r_{sdf_{min}} r_{sdf_{max}}}{r_{sdf_{max}} - r_{sdf_{min}}}$ is a normalization factor ensuring $\mathcal{F}_{sdf_{i}}$ is continuous, and $d_{obs_{i}}$ is the distance between the UAV and the nearest obstacle. Both $r_{sdf_{min}}$ and $r_{sdf_{max}}$ are listed in Table \ref{table:parameters_table}. This inverse-distance formulation reflects the increasing risk as the UAV approaches obstacles, with the cost growing rapidly in their vicinity. The non-insertion cost function $\mathcal{F}_{ch_{i}}$ is formulated as follows:
\begin{equation}
    \label{eq:convex_hull_cost}
    \mathcal{F}_{ch_{i}} =
    \begin{cases}
        0 & d_{ch_{i}} \geq r_{ch_{max}} \\
        1 - \frac{d_{ch_{i}}}{r_{ch_{max}}} & 0 < d_{ch_{i}} < r_{ch_{max}} \\
        1 & d_{ch_{i}} \leq 0
    \end{cases}
\end{equation}
where $d_{ch_{i}}$ is the distance to the nearest convex hull and $r_{ch_{max}}$ is the maximum influence radius of convex hulls. Under high wind, communication, or localization risks, the algorithm prioritizes safety by selecting trajectories that minimize this cost function over time or energy efficiency.

\subsubsection{Energy cost}
\label{sec:energy_cost}
To evaluate the energy consumption of a planned trajectory, this approach integrates physics-based principles with experimental data. Authors in \cite{guoku2022} proposed a model for quadrotor UAVs with BLDC motors, showing that energy consumption varies with the flight state. Based on this model, it can be hypothesized that horizontal movements at constant speed generally require more power than hovering, as the UAV must tilt to generate horizontal thrust while still maintaining altitude and overcoming aerodynamic drag. Vertical movements further affect this consumption. The pitch/roll power relationship with the vertical axis forms a quadric surface \cite{hilbert1999}, described by a general equation:
\begin{equation}
    \label{eq:quadric_surface}
    \begin{aligned}
        {} & ax^{2}+by^{2}+cz^{2}+dxy+eyz+fxz\\
        & \ \ \ +gx+hy+kz+L = 0 
    \end{aligned}
\end{equation}
For a balanced quadrotor with symmetrical propulsion, the power required for steady flight is invariant under reflections of the body roll and pitch axes (i.e., $P(x, y, z) = P(-x, y, z)$ and $P(x, y, z) = P(x, -y, z)$). The mixed terms $dxy$, $eyz$, and $fxz$ change sign under these reflections and therefore must disappear, making $d=e=f=0$. Physically, these terms would represent coupling between roll, pitch, and vertical motion arising from structural or aerodynamic asymmetries, which are neglected in the current model. Since the quadric is defined only up to an arbitrary scale factor, $L$ is fixed to unity to remove the scale ambiguity. The resulting model contains six unknown coefficients $(a,b,c,g,h,k)$, which can be uniquely identified from steady-state power measurements collected in the six principal flight directions ($\pm Z, \pm Roll, \pm Pitch$). To locate a point on this surface during optimization, a projection is made along a unit vector $\hat{v} = (v_{x}, v_{y}, v_{z})$, parameterized by $t$: $x = t v_{x}, y = t v_{y}, z = t v_{z}$. Substituting into Eq. \ref{eq:quadric_surface} simplifies to:
\begin{equation}
    \label{eq:simplified_quadric_equation}
    At^{2} + Bt + L = 0
\end{equation}
where $A = av_{x}^{2} + bv_{y}^{2} + cv_{z}^{2}$, $B = gv_{x} + hv_{y} + kv_{z}$, and $L = 1$. Solving this equation using the quadratic formula yields points on the quadric surface along the unit vector, providing steady-state power consumption $P(\hat{v})$ at these coordinates. 

The transient state is modeled as the change in kinetic energy between two planned velocity points. This simplified model captures the energy influence of short-duration maneuvers or bursts of acceleration not reflected in the steady-state model. The energy consumption cost function is given by:
\begin{equation}
    \label{eq:energy_cost}
    \mathcal{F}_{Energy} = \sum_{i=1}^{Q} \big( P(\hat{v}_{i})\,\Delta t
    + \frac{m_{UAV}}{2} \left| \left\|\vec{v}_{i}\right\|^{2} - \left\|\vec{v}_{i-1}\right\|^{2} \right| \big)
\end{equation}
This model generalizes to any velocity-based trajectory representation. Unlike the time cost function, the energy cost function includes a directional factor influencing the optimizer. In low battery conditions, the algorithm prioritizes energy and time over safety.
\subsection{Algorithm}
\label{sec:risk_aware_path_planning_solver}

The adaptive risk-aware path planning algorithm consists of two steps. First, a multi-objective solver returns an approximate Pareto front of feasible trajectories. Then, these trajectories are filtered using a voting algorithm to select the most suitable one based on real-time mission risks.

\subsubsection{Solver}
\label{sec:solver}

The approach promotes exploration of the search space without bias toward any specific behavior. To achieve this, the optimizer must remain impartial and retain valuable solutions that minimize individual objectives or a combination of them (Fig. \ref{fig:arena_pipeline}, block (3)). Among several multi-objective optimizers that satisfy these requirements, the NSGA-II genetic algorithm was selected for its efficiency with continuous problems \cite{deb2002}. It uses polynomial mutation and simulated binary crossover to explore the search space. In the default implementation, NSGA-II initializes a population of $\mu$ individuals using a normal distribution within the bounded search space. However, due to hard constraints in the problem definition, this initialization method mostly generates infeasible solutions, wasting computation time and increasing the risk of premature convergence. To address this, the population is initialized using the Rapidly-exploring Random Tree (RRT) algorithm \cite{lavalle1998rrt}, which efficiently explores large, high-dimensional, and complex state spaces while satisfying hard constraints (Fig. \ref{fig:arena_pipeline}, block (2)). This initialization ensures a set of feasible trajectories and provides sparse coverage of the environment without introducing behavioral bias. The optimization then yields a set of feasible solutions forming a Pareto front.

\subsubsection{Voting}
\label{sec:voting}
A novel voting algorithm is implemented to select the trajectory to execute from the approximated Pareto set according to externally estimated operational risk indicators (Fig.~\ref{fig:arena_pipeline}, block (4)), leveraging the plurality of optimized solutions to fit the current mission status.

Let $[\mathcal{WR}, \mathcal{CR}, \mathcal{LR}, \mathcal{BR}] \in [0, 1]^{4}$ denote normalized estimates of wind, communication, localization, and battery operational risks, respectively. A value of 0 corresponds to nominal operational risk, while 1 denotes a critical operating condition. These quantities are assumed to be provided by an external module. As mentioned in \cite{petit2024}, their computation is application- and platform-dependent and is therefore outside the scope of this work. ARENA treats them as exogenous inputs.

The costs $f_{n,e}$ of every solution $n \in N$ and objective $e \in \mathcal{F}$ are first normalized over the generated Pareto set.
\begin{equation}
    \label{eq:cost_normalization}
    \hat{f}_{n,e} = \frac{f_{n,e} - \min_{n_{l} \in \mathcal{N}} f_{n_{l},e}}{\max_{n_{l} \in \mathcal{N}} f_{n_{l},e} - \min_{n_{l} \in \mathcal{N}} f_{n_{l},e}} \in [0, 1]
\end{equation}
In this paper, the influence of each risk on each objective is described by the matrix $\mathbf{A} = [a_{e,r}]$, with rows $Time$, $Safety$, $Energy$ $\{T, S, E\}$ and columns $\mathcal{WR}$, $\mathcal{CR}$, $\mathcal{LR}$, $\mathcal{BR}$:
\begin{equation}
    \label{eq:influence_matrix}
    \mathbf{A} =
    \begin{bmatrix}
        -1 & -1 & -1 & 1 \\
        1 & 1 & 1 & -1 \\
        \tfrac{1}{2} & 0 & 0 & \tfrac{1}{2}
    \end{bmatrix}
\end{equation}
Each column is scaled to a largest entry of 1, so that every risk has the same authority. Each coefficient takes the normalized product form of state variable weights \cite{liu1997}, here driven by the external risks through $\mathbf{A}$ rather than by each objective's own value:
\begin{equation}
    \label{eq:coefficient_adjustment_functions}
    k_{e} = \frac{k_{e,0} \prod_{r} (1 - r)^{-\beta a_{e,r}}}{\sum_{m \in \{T, S, E\}} k_{m,0} \prod_{r} (1 - r)^{-\beta a_{m,r}}}
\end{equation}
where $k_{e,0}$ are the nominal coefficients and $\beta$ is a gain. By construction, $k_{T} + k_{S} + k_{E} = 1$ for any risks and $k_{e} = k_{e,0}$ under nominal conditions. For low risks, Eq.~\ref{eq:coefficient_adjustment_functions} reduces to a linear adjustment of the nominal coefficients,
\begin{equation}
    \label{eq:coefficient_first_order}
    k_{e} \approx k_{e,0} \Big(1 + \beta \big(\textstyle\sum_{r} a_{e,r} r - \sum_{m} k_{m,0} \sum_{r} a_{m,r} r\big)\Big)
\end{equation}
while, as a risk approaches 1, the coefficient of the objective it influences most tends to 1.

The selected trajectory minimizes the augmented weighted Tchebycheff scalarization \cite{steuer1983, wilde2024} of the normalized costs:
\begin{equation}
    \label{eq:voting_equation}
    i^{*} = \operatorname*{arg\,min}_{n \in \mathcal{N}} \Big( \max_{e} k_{e} \hat{f}_{n,e} + \epsilon \sum_{e} k_{e} \hat{f}_{n,e} \Big)
\end{equation}
where $\epsilon \ll 1$ breaks ties toward non-dominated solutions. Unlike weighted sums of costs or of ranks, which can only select part of the Pareto set, every Pareto-optimal solution is the unique minimizer of Eq.~\ref{eq:voting_equation} for some coefficients \cite{steuer1983, miettinen2012}, so any trajectory of $\mathcal{N}$ can be reached by the risks. We use $k_{e,0} = 1/3$, $\beta = 2$, and $\epsilon = 10^{-3}$. The entries of $\mathbf{A}$ are specific to the power line inspection problem studied here and can be adapted to other UAV platforms and missions.

\subsubsection{Parameters}
\label{sec:hyperparameters}

Table \ref{table:parameters_table} gives an overview of the needed UAV parameters and the most important hyperparameters of the path planning method. Note that the NSGA-II algorithm's crossover and mutation parameters were chosen according to the recommendation of the Pagmo library's NSGA-II implementation \cite{Biscani2020}. For tuning, the authors suggest starting with $N_{pop} = 40$ and $N_{gen} = 300$, and then adjusting $\delta_{RRT}$ and $N_{nurbs}$ to ensure sufficient control points and resolution for the optimization problem.

%
\begin{table}[h]
    \caption{ARENA hyperparameters}
    \label{table:parameters_table}

    \resizebox{\columnwidth}{!}
    {
        \renewcommand{\arraystretch}{1.5}
        \begin{tabular}{c|c|c|c|}
             \cline{2-4}
             & \textbf{Hyperparameter} & \textbf{Symbol} & \textbf{Value} \\ \hline
             \multicolumn{1}{|c|}{General}
             & \begin{tabular}[c]{@{}l@{}}
                UAV mass ($kg$) \\ UAV speed ($m/s$) \\ UAV acceleration ($m/s^{2}$) \\
                NURBS curve degree \\ SDF influence radius \\ Convex hull influence radius
            \end{tabular}
            & \begin{tabular}[c]{@{}c@{}}
                $m_{uav}$ \\ $v_{max}$ \\ $a_{max}$ \\
               $p$ \\ $[ r_{sdf_{min}}, r_{sdf_{max}} ]$ \\ $r_{ch_{max}}$
            \end{tabular}
            & \begin{tabular}[c]{@{}c@{}}
                $\mathbb{R}_{> 0}$ \\ $\mathbb{R}$ \\ $\mathbb{R}$ \\
               $\{ x \in \mathbb{N} \mid 1 < x \leq 5 \}$ \\ $\mathbb{N}_{> 0}$ \\ $\mathbb{N}_{> 0}$
            \end{tabular} \\ \hline
            
             \multicolumn{4}{c}{} \\ \hline

            \multicolumn{1}{|c|}{\textbf{Algorithm step}} & \textbf{Hyperparameter} & \textbf{Symbol} & \textbf{Value} \\ \hline
            \multicolumn{1}{|c|}{Initialization} & RRT node distance & $\delta_{RRT}$ & $\mathbb{R}_{> 0}$ \\ \hline
            \multicolumn{1}{|c|}{NSGA-II}
            & \begin{tabular}[c]{@{}l@{}}
                Number of generations\\ Population size\\ NURBS sampling size
            \end{tabular}
            & \begin{tabular}[c]{@{}c@{}}
                $N_{gen}$\\ $N_{pop}$\\ $N_{nurbs}$
            \end{tabular}
            & \begin{tabular}[c]{@{}c@{}}
                $\mathbb{N}_{> 0}$\\ $\mathbb{N}_{> 0}$\\ $\mathbb{N}_{> 0}$
            \end{tabular} \\ \hline
        \end{tabular}
    }
\end{table}
\vspace*{-\baselineskip}

\section{Results}
\label{sec:results}

This paper evaluates the performance of ARENA in two simulated environments and further confirms its behavior and results in a real-world scenario. The experiments are designed to answer the following questions:

\begin{itemize}
    \item [\textbf{Q1.}] How effectively does ARENA adapt its behavior to varying mission risk profiles (e.g., battery level, wind, GPS uncertainty)?
    
    \item [\textbf{Q2.}] Can the planner consistently generate an approximate Pareto front of trajectories that balance safety, energy consumption, and path duration across heterogeneous scenarios?
    
    \item [\textbf{Q3.}] In real-world settings, can the planner produce feasible and risk-aware trajectories that preserve the qualitative behaviors observed in simulation (e.g., obstacle-averse or energy-efficient paths)?

    \item [\textbf{Q4.}] How accurate is the proposed power consumption model when applied to real flight data, and is it sufficiently precise to guide energy-aware trajectory optimization?
\end{itemize}

\subsection{Simulated experiments}
\label{sec:simulated_experiments}

To answer these questions, the planner is applied to the LineDrone \cite{miralles2018}, a power line inspection UAV \cite{wenkai2017, park2019} developed by Hydro-Québec and DroneVolt for in-contact non-destructive inspections of energized high-voltage power lines. It carries a dual-antenna GPS for localization and orientation, a vertically mounted LiDAR for obstacle detection, and a Jetson Xavier NX for onboard computation. The algorithm is implemented in C++11 and deployed with the Robot Operating System (ROS2), using the Open Motion Planning Library (OMPL) \cite{ioan2012} for trajectory initialization, OctoMap \cite{hornung2013} for obstacle detection, and Pagmo2 \cite{Biscani2020} for multi-objective optimization. Simulations were performed on an 8-core 2.3\,GHz i7 CPU with 16\,GB RAM, with the simulated drone mirroring the LineDrone's perception and localization capabilities.

\label{sec:simulated_experiments2}

\begin{figure}[]
    \centering
    \includegraphics[width=0.98\linewidth]{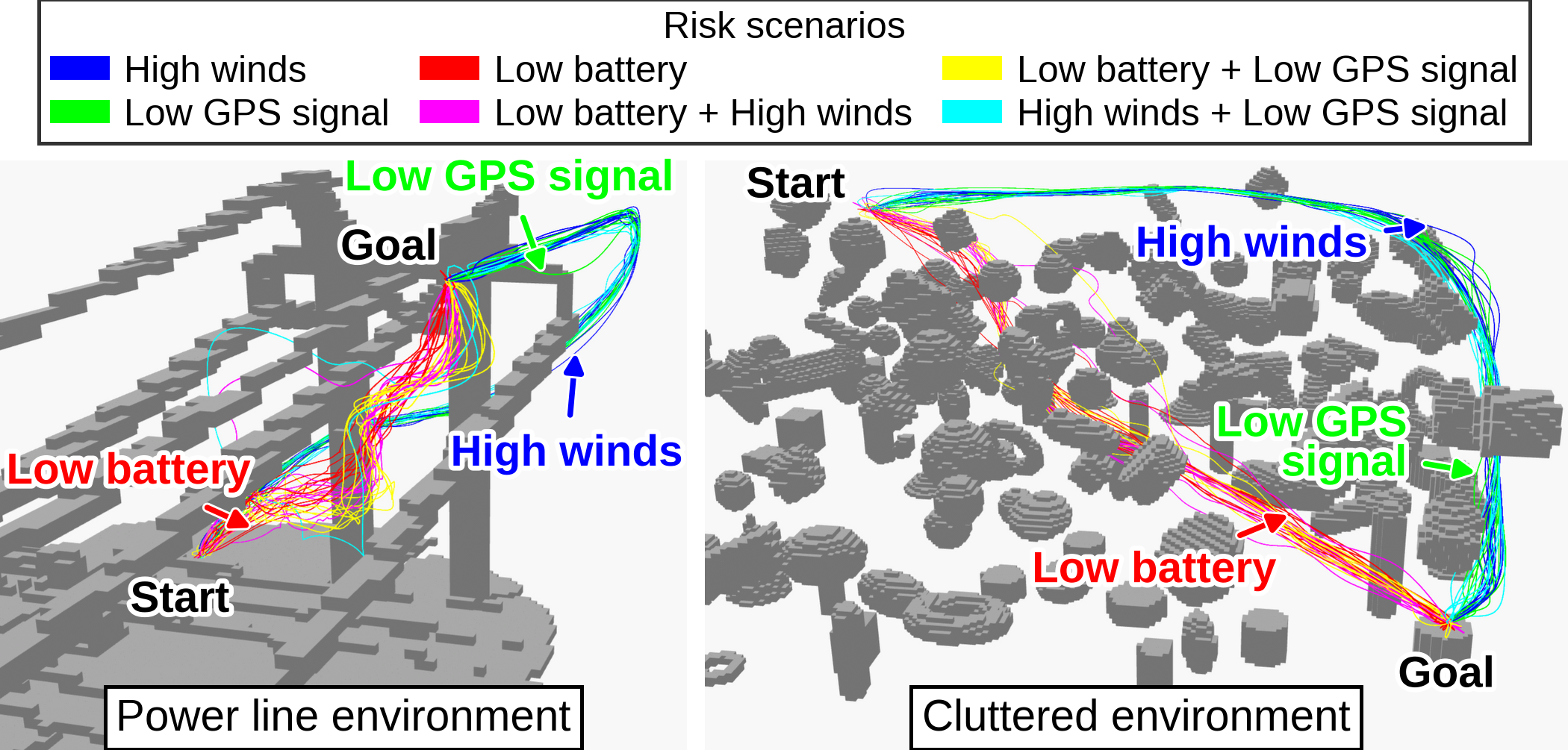}
    \captionsetup{width=\linewidth}
    \caption{Path comparison for different operational risk scenarios in two simulated environments, both with trajectories preferred in different operational risk scenarios. Gray voxels denote obstacles generated by OctoMap.}
    \label{fig:path_comparison}
    \vspace*{-\baselineskip}
\end{figure}
To answer \textbf{Q1} and \textbf{Q2}, ARENA was evaluated in multiple simulated environments, each presenting unique challenges: 6 reconstructed high-voltage power lines scenarios, a high-rise urban construction site, and randomly generated cluttered environments. We focus here on the 2 environments of Fig. \ref{fig:path_comparison}, where the shown trajectories exhibit distinct behaviors: (a) under high battery risk, red trajectories are faster but offer less obstacle clearance, (b) under high wind and/or GPS risk, green and cyan trajectories keep greater obstacle distances, (c) blue trajectories save energy while maintaining a minimum obstacle distance, and (d) intermediate behaviors arise from mixed operational risk combinations.

To benchmark the algorithm's ability to consistently generate a good approximated Pareto front that balances objectives across heterogeneous scenarios (\textbf{Q2}), we ran the algorithm with over-designed hyperparameters ($\delta_{RRT} = 5.0$, $N_{gen} = 5000$, $N_{pop} = 252 $, $N_{nurbs} = 250$) 500 times in each environment. We then reconstructed a high-fidelity Pareto reference front from the 126k optimized trajectories.

\begin{figure}[]
    \centering
    \includegraphics[width=0.96\linewidth]{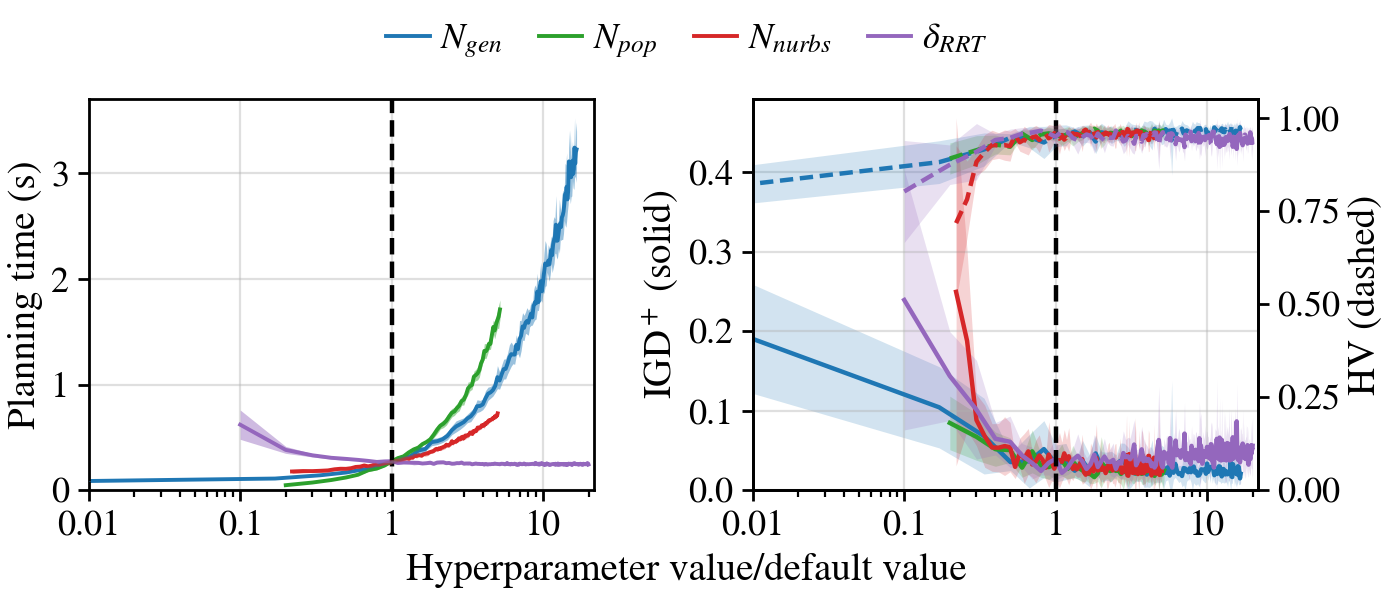}
    \captionsetup{width=\linewidth}
    \caption{Computation time and Pareto front quality for different hyperparameter settings in the power lines environment. Default values are $N_{gen} = 300$, $N_{pop} = 40$, $N_{nurbs} = 50$, and $\delta_{RRT} = 10.0$.}
    \label{fig:computation_analysis}
    \vspace*{-\baselineskip}
\end{figure}

The execution time and the quality of the generated fronts depend on several hyperparameters. To analyze these effects, ARENA's most important hyperparameters were swept individually, with the others fixed at their default values (Fig. \ref{fig:computation_analysis}). Front quality was evaluated with the hypervolume (HV) and inverted generational distance plus (IGD$^+$) metrics, relative to the high-fidelity reference front. $N_{pop}$ and $N_{gen}$ have the largest impact on planning time, while $N_{nurbs}$ and $\delta_{RRT}$ most affect front quality. The results show that the selected default values offer a good trade-off between computational efficiency and solution quality in simulation (\textbf{Q2}).

ARENA, with default hyperparameters, was also compared to state-of-the-art planners in Table \ref{tab:baselines}. All planners' outputs were evaluated at high resolution to accurately represent the results of discretized planners. In both environments, ARENA outperforms all constraint-aware baselines in best path duration, best energy, and overall Pareto front quality. MOAR-3D kino.\ sweep achieves a higher best mean clearance only because, with $k_{S} = 1.0$, nothing drives it towards the goal, so it wanders in the zero-cost zone. Since ARENA's signed distance field has a maximum influence radius of 9\,m in these scenarios, beyond which clearance no longer reduces the safety cost, ARENA's results are considered to reach the optimal cost value. Notably, no baseline outperformed the reference front, confirming its high-fidelity (\textbf{Q2}).
\begin{table*}[t]
\centering
\caption{Comparison with baselines (20 runs per non-deterministic planner)}
\label{tab:baselines}
\scriptsize
\setlength{\tabcolsep}{1.2pt}
\begin{tabular*}{\textwidth}{@{\extracolsep{\fill}}l@{\hspace{10pt}}cccccccccccccccc@{}}
\toprule
 & \multicolumn{8}{c}{Cluttered environment} & \multicolumn{8}{c}{Power lines environment} \\
\cmidrule(lr){2-9} \cmidrule(lr){10-17}
 &  &  & \multicolumn{4}{c}{Best of the front} &  &  &  &  & \multicolumn{4}{c}{Best of the front} &  &  \\
\cmidrule(lr){4-7} \cmidrule(lr){12-15}
 & SR & Pl.\,Time & Dur. & Energy & $d_\text{min}$ & $d_\text{mean}$ & HV & IGD$^+$ & SR & Pl.\,Time & Dur. & Energy & $d_\text{min}$ & $d_\text{mean}$ & HV & IGD$^+$ \\
Planner & (\%) & (s) & (s) & (kJ) & (m) & (m) &  &  & (\%) & (s) & (s) & (kJ) & (m) & (m) &  &  \\
\midrule
ARENA (ours) & \textbf{100} & 0.31\sd{$\pm$.01} & \textbf{254}\sd{$\pm$4} & \textbf{802}\sd{$\pm$13} & 7.0\sd{$\pm$.2} & 9.3\sd{$\pm$.5} & \textbf{0.93}\sd{$\pm$.02} & \textbf{0.05}\sd{$\pm$.02} & \textbf{100} & 0.27\sd{$\pm$.02} & \textbf{151}\sd{$\pm$9} & \textbf{495}\sd{$\pm$30} & 7.3\sd{$\pm$.2} & 10.1\sd{$\pm$1.6} & \textbf{0.92}\sd{$\pm$.04} & \textbf{0.05}\sd{$\pm$.03} \\
MOAR-3D kino.\ sweep & \textbf{100} & 15.4$^\dagger$\sd{(4.5)} & 338 & 1072 & \textbf{7.1} & \textbf{11.3} & 0.84 & 0.13 & \textbf{100} & 19.5$^\dagger$\sd{(17.0)} & 165 & 540 & \textbf{7.5} & \textbf{10.6} & 0.83 & 0.17 \\
Fast-Planner~\cite{zhou2019} & \textbf{100} & \textbf{0.028}\sd{$\pm$.002} & 339 & 1075 & 1.3 & 6.0 & 0.17 & 0.59 & \textbf{100} & \textbf{0.012} & 170 & 558 & 1.8 & 6.5 & 0.18 & 0.83 \\
Kino.\ RRT$^*$ & 90 & 60.0 & 369\sd{$\pm$8} & 1171\sd{$\pm$26} & 1.6\sd{$\pm$.5} & 6.2\sd{$\pm$.6} & 0.25\sd{$\pm$.10} & 0.51\sd{$\pm$.09} & \textbf{100} & 60.0 & 206\sd{$\pm$5} & 672\sd{$\pm$15} & 2.7\sd{$\pm$.9} & 6.5\sd{$\pm$.5} & 0.28\sd{$\pm$.16} & 0.69\sd{$\pm$.21} \\
\cmidrule(lr){1-17}
\rx{A$^*$} & \rx{0} & \rx{0.017\sd{$\pm$.002}} & \rx{251} & \rx{796} & \rx{1.5} & \rx{5.5} & -- & -- & \rx{0} & \rx{0.004\sd{$\pm$.003}} & \rx{139} & \rx{458} & \rx{2.0} & \rx{6.3} & -- & -- \\
\rx{RRT$^*$} & \rx{0} & \rx{0.91\sd{$\pm$.01}} & \rx{270\sd{$\pm$4}} & \rx{854\sd{$\pm$14}} & \rx{1.6\sd{$\pm$.4}} & \rx{5.6\sd{$\pm$.5}} & -- & -- & \rx{0} & \rx{0.75} & \rx{186\sd{$\pm$22}} & \rx{605\sd{$\pm$69}} & \rx{2.3\sd{$\pm$.9}} & \rx{6.4\sd{$\pm$.8}} & -- & -- \\
\rx{MOAR-3D~\cite{petit2024} sweep} & \rx{0} & \rx{0.36$^\dagger$\sd{(.60)}} & \rx{255} & \rx{808} & \rx{7.1} & \rx{9.6} & \rx{0.98} & \rx{0.01} & \rx{0} & \rx{0.22$^\dagger$\sd{(.25)}} & \rx{147} & \rx{484} & \rx{7.5} & \rx{9.8} & \rx{0.95} & \rx{0.03} \\
\rx{Ahmed and Deb~\cite{ahmed2011}} & \rx{0} & \rx{1.51\sd{$\pm$.08}} & \rx{248\sd{$\pm$6}} & \rx{783\sd{$\pm$19}} & \rx{4.3\sd{$\pm$1.1}} & \rx{8.0\sd{$\pm$.8}} & \rx{0.79\sd{$\pm$.13}} & \rx{0.10\sd{$\pm$.09}} & \rx{0} & \rx{0.94\sd{$\pm$.04}} & \rx{152\sd{$\pm$11}} & \rx{496\sd{$\pm$34}} & \rx{6.5\sd{$\pm$1.2}} & \rx{10.4\sd{$\pm$1.3}} & \rx{0.85\sd{$\pm$.07}} & \rx{0.14\sd{$\pm$.07}} \\
\rx{A$^*$pex~\cite{zhang2022}} & \rx{0} & \rx{23.9\sd{$\pm$.5}} & \rx{255} & \rx{808} & \rx{7.1} & \rx{11.9} & \rx{0.98} & \rx{0.02} & \rx{0} & \rx{103\sd{$\pm$2}} & \rx{147} & \rx{484} & \rx{7.5} & \rx{12.4} & \rx{0.94} & \rx{0.03} \\
\rx{PO-RRT$^*$~\cite{batonisashvili2026}} & \rx{0} & \rx{5.01} & \rx{311\sd{$\pm$13}} & \rx{986\sd{$\pm$41}} & \rx{1.9\sd{$\pm$.4}} & \rx{6.3\sd{$\pm$.8}} & \rx{0.35\sd{$\pm$.11}} & \rx{0.41\sd{$\pm$.10}} & \rx{0} & \rx{5.01} & \rx{218\sd{$\pm$46}} & \rx{707\sd{$\pm$143}} & \rx{4.2\sd{$\pm$1.2}} & \rx{8.7\sd{$\pm$.9}} & \rx{0.48\sd{$\pm$.17}} & \rx{0.47\sd{$\pm$.23}} \\
\bottomrule
\end{tabular*}
\\[2pt]
\parbox{\textwidth}{\fontsize{6}{7}\selectfont SR: success rate. Pl.\,Time: planning time. Dur.: duration. $d_\text{min}$, $d_\text{mean}$: minimum and mean clearance. HV: hypervolume. IGD$^+$: inverted generational distance plus. \rx{Gray: trajectories violate constraints.} $^\dagger$median (IQR) shown instead of mean $\pm$ std because of skewed distribution.}
\vspace*{-1.5\baselineskip}
\end{table*}

\begin{figure}[]
    \centering
    \includegraphics[width=0.96\linewidth]{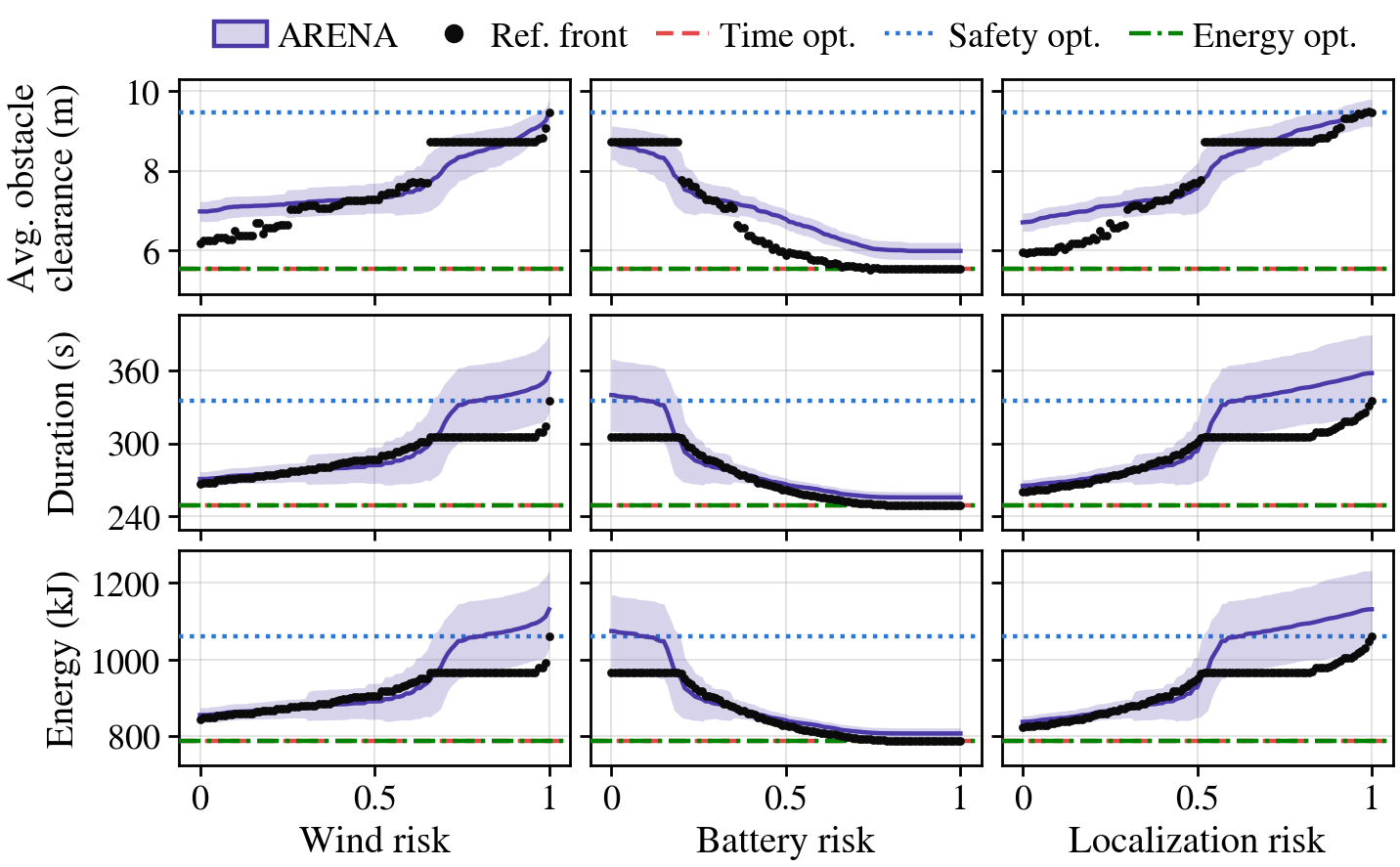}
    \captionsetup{width=\linewidth}
    \caption{Metric sensitivity to risk factors in the cluttered environment. The ARENA curve shows the mean and std of 50 planning requests using default hyperparameters.}
    \label{fig:risks_variation_both_metrics_all_risks}
    \vspace*{-\baselineskip}
\end{figure}

To evaluate the decision mechanism independently, the normalized risk variables were manually swept from 0 to 1 in increments of 0.1 on the high-fidelity reference front of the cluttered environment and on 50 planning requests of ARENA at default hyperparameters on every increment. This controlled study isolates the sensitivity of the planner to each risk source. Fig. \ref{fig:risks_variation_both_metrics_all_risks} shows the average distance to obstacles, path duration, and total energy consumption for Pareto front trajectories, analyzed as functions of 3 of the 4 operational risks mentioned in Section \ref{sec:voting}: low battery, high winds, and low localization precision. Low communication operational risk was excluded as it mirrors the low localization precision operational risk in Eq. \ref{eq:influence_matrix}. Results indicate that with high battery operational risk, the planner prioritizes trajectories with lower path duration and energy consumption. With high wind or low localization precision, the planner prioritizes trajectories with higher obstacle distances. Using the reference front, the planner's trajectory selection spans 100\% of the benchmarks' ranges for all metrics. With ARENA with default hyperarameters, the selection spans no less than $88.4\% \pm 5.5\%$. These results demonstrate the algorithm's ability to adapt its decision-making process based on risk inputs. Similar results across all simulated environments confirmed the algorithm's adaptive weighting based on varying risks (\textbf{Q1}).

\subsection{Physical Experiments}
\label{sec:physical_experiment}

\begin{figure}[]
    \begin{minipage}{\linewidth}
        \centering
        \captionsetup[subfloat]{textfont=small}
        \subfloat[]{
            \includegraphics[width=0.49\linewidth, height=1.25in]{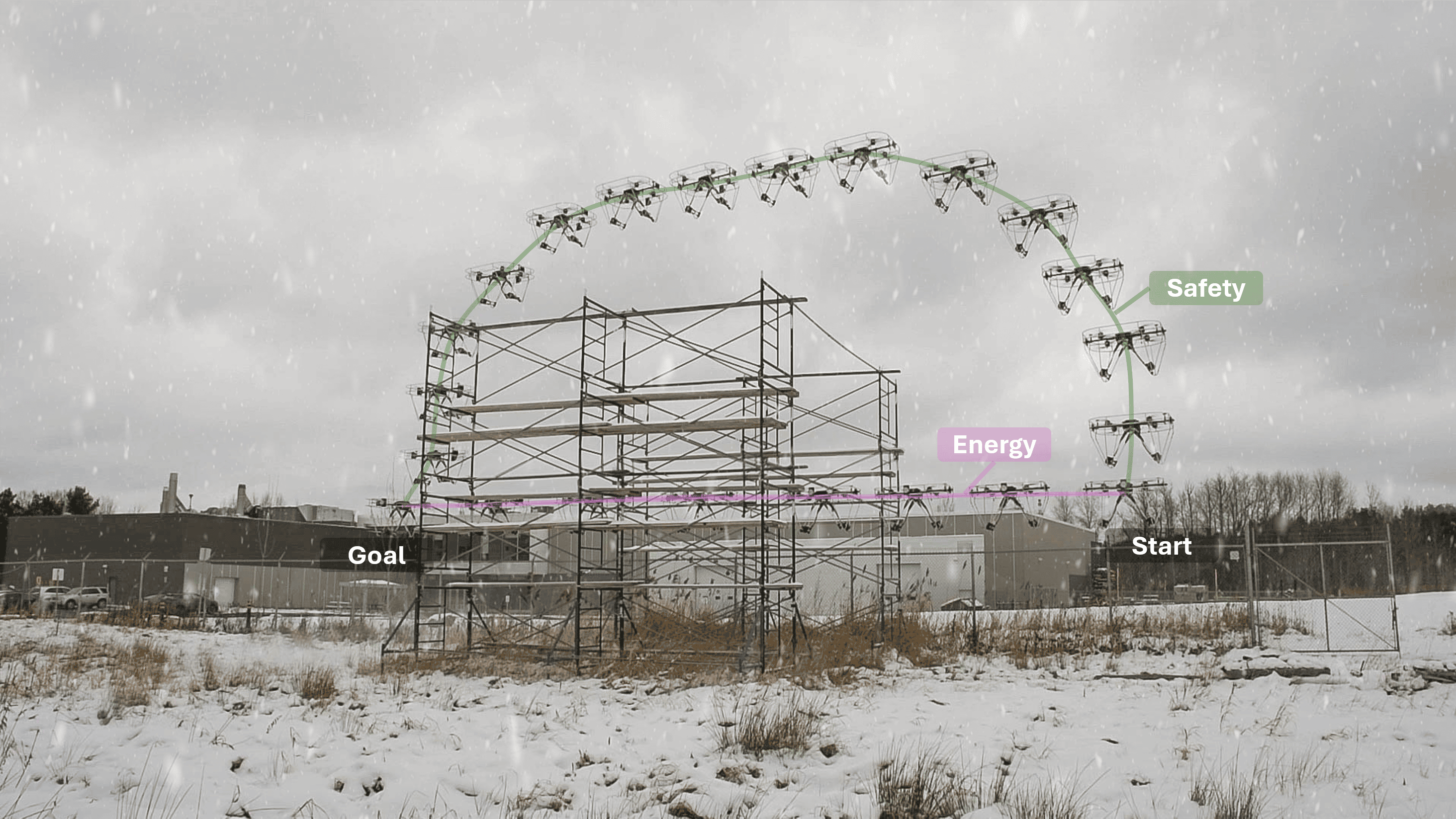}
            \label{fig:drone_outside_mashup_pic}
        }
        \subfloat[]{
            \includegraphics[width=0.48\linewidth, height=1.25in]{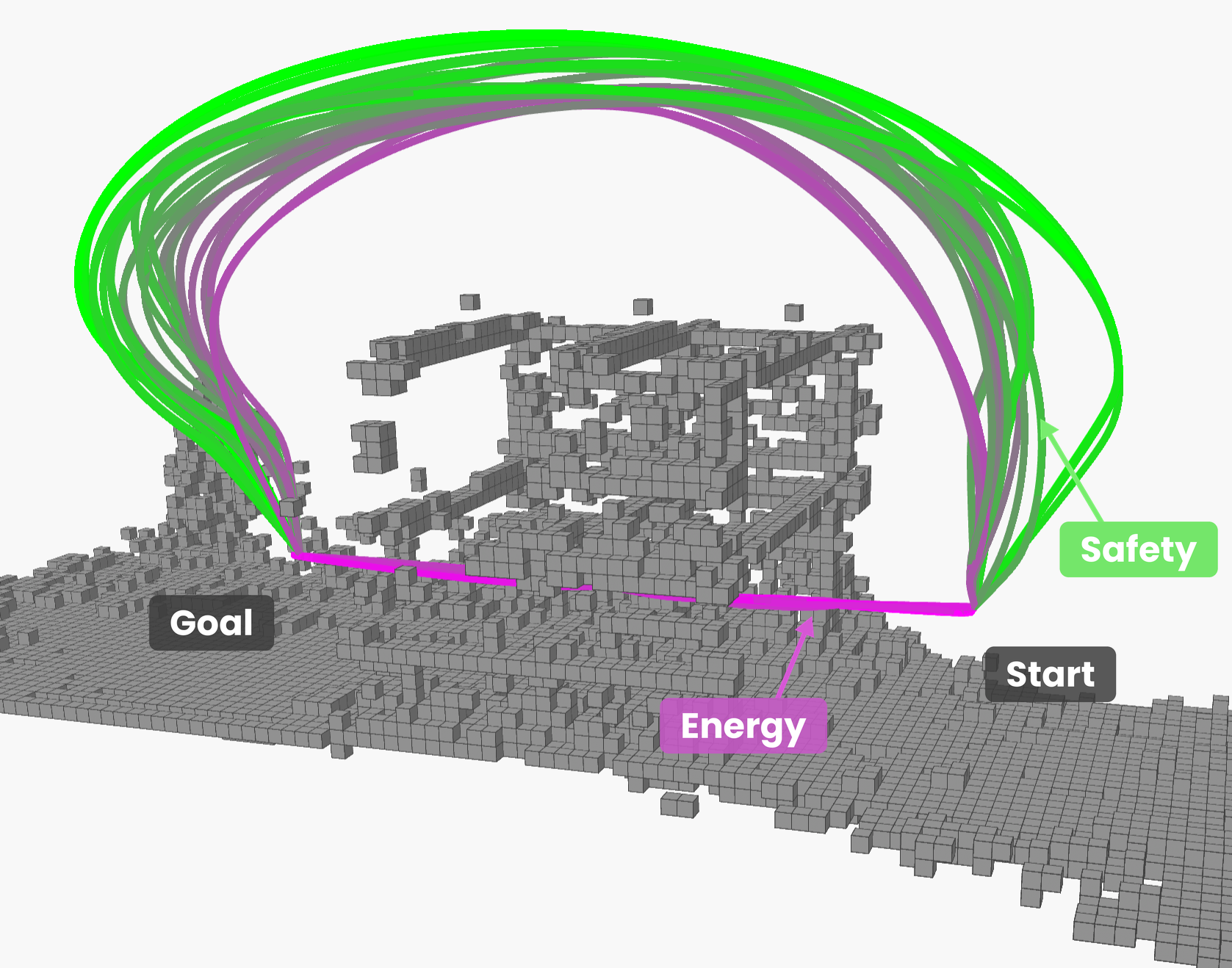}
            \label{fig:solution_set_echafaudage}
        } 
        \captionsetup{width=\linewidth}
        \caption{Flight test with the LineDrone on a power lines model. (a) Path reconstructed from a video. (b) Planned Pareto front by ARENA. Safety trajectory has coefficients $k_{s}=0.7$, $k_{t}=0.1$, and $k_{e}=0.2$. Energy trajectory has coefficients $k_{s}=0.05$, $k_{t}=0.05$, and $k_{e}=0.9$.}
        \label{fig:flight_test}
    \end{minipage}
    \vspace*{-\baselineskip}
\end{figure}

To answer \textbf{Q3}, the approach was validated through real-world flight tests with the LineDrone on a scaffold structure mimicking a power line inspection mission. As shown in Fig. \ref{fig:flight_test}, the UAV followed two trajectories: an energy-efficient one, passing between structures to reach the goal faster, and a safety-oriented one, keeping a larger average distance to obstacles. With the default hyperparameters, computation times on hardware similar to the LineDrone's onboard computer were slightly below 1\,s. These results show that, in a real-world scenario, the algorithm plans behaviors tailored to specific risks, prioritizing either energy savings or clearance.

\begin{figure*}[]
    \centering
    \subfloat[]{
        \includegraphics[width=2.15in]{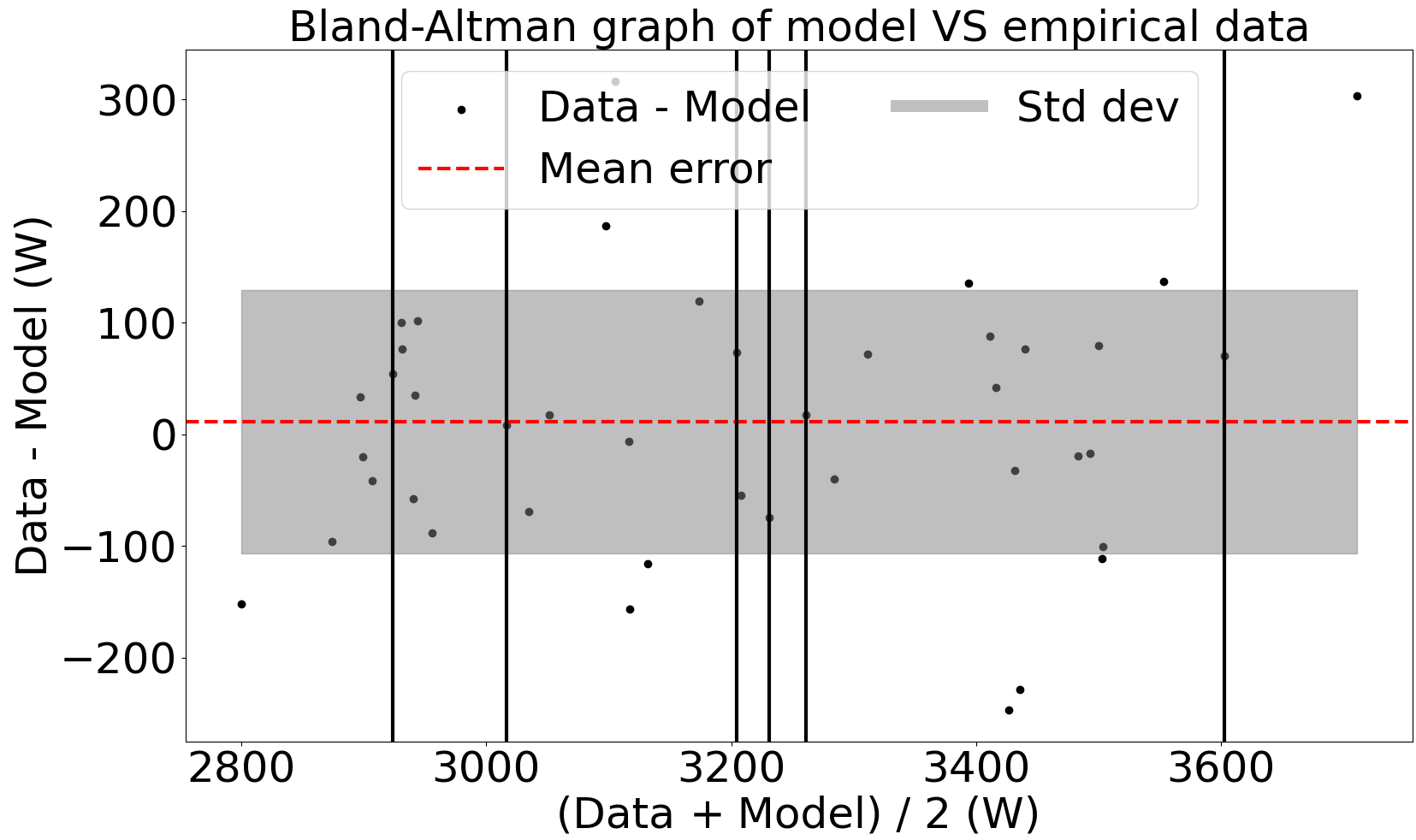}
        \label{fig:bland_altman_plot_more_points}
    }
    \hspace{0.025em}%
    \subfloat[]{
        \includegraphics[width=2.15in]{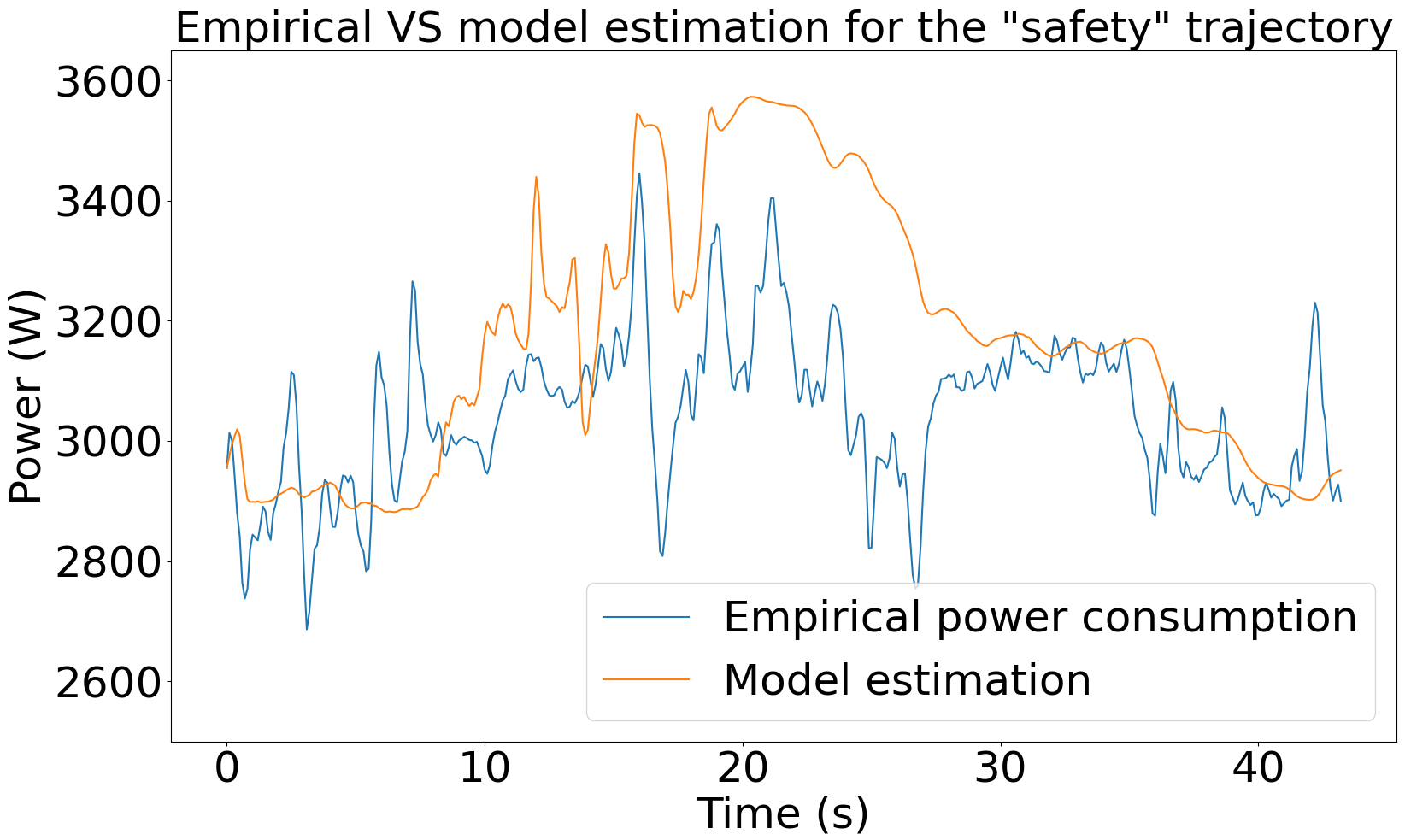}
        \label{fig:empirical_power_consumption_vs_model_estimation_for_safety_traj}
    }
    \hspace{0.025em}%
    \subfloat[]{
        \includegraphics[width=2.15in]{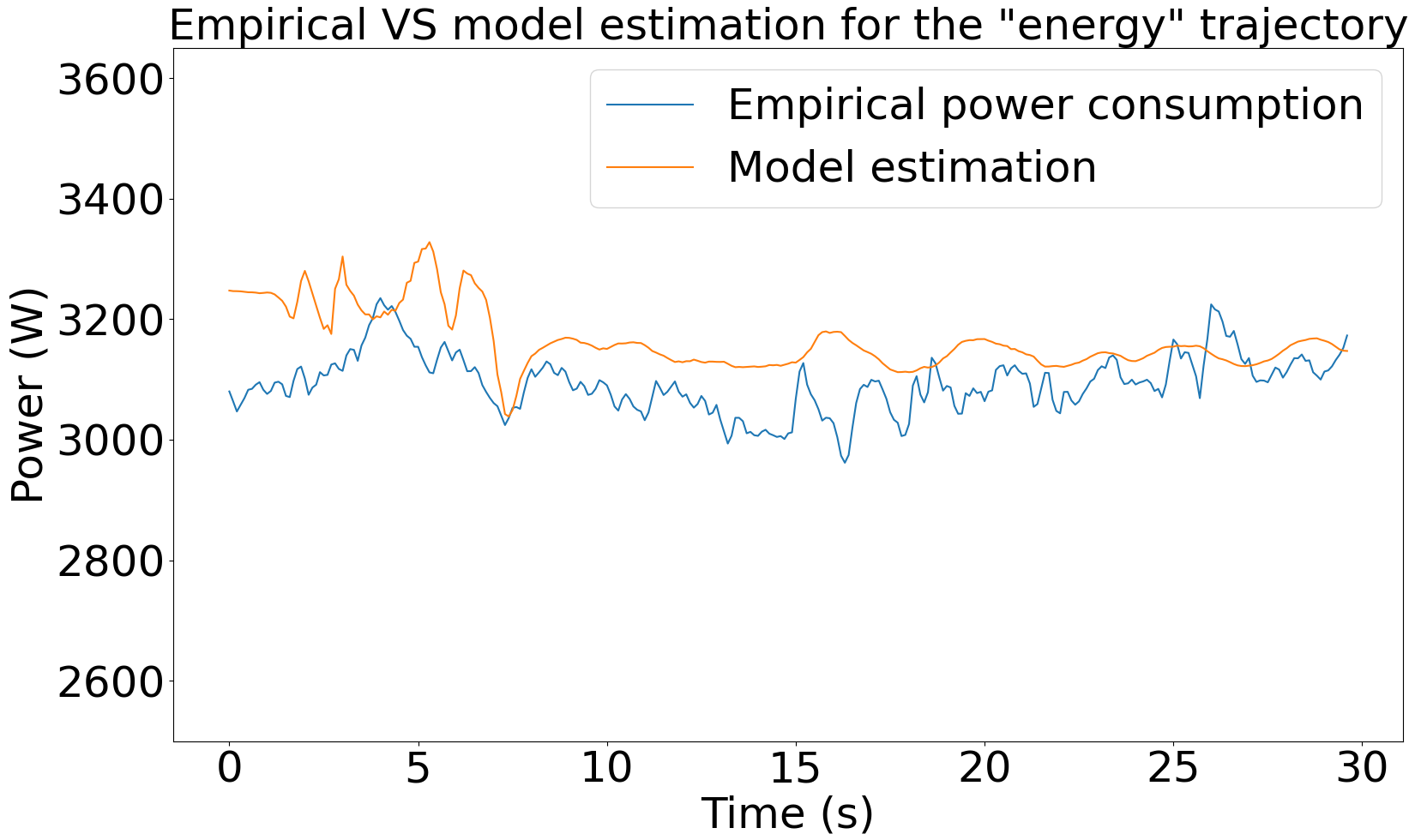}
        \label{fig:empirical_power_consumption_vs_model_estimation_for_energy_traj}
    }
    \captionsetup{width=\linewidth}
    \caption{Empirical power consumption data of a flight test with the LineDrone in windy conditions. (a) Bland-Altman graph of the empirical data VS the power model consumption's best-fitting model (b)-(c) Model estimation VS real power consumption for the "safety" and "energy" trajectories in Fig. \ref{fig:flight_test}. Note that the mismatch between estimation and real-flight data in (b) at 20 seconds is due to a substantial wind gust.}
    \label{fig:power_consumption_model_visualization}
    \vspace*{-\baselineskip}
\end{figure*}

To construct and validate the power consumption model, 42 real-world flight tests were conducted, where the UAV flew in 42 different directions at a speed of 2 m/s for 14 seconds to capture steady-state power consumption. Data from the six directional axes (14\% of the dataset) were used to construct the model, reserving the remaining 86\% for validation. Fig. \ref{fig:bland_altman_plot_more_points} illustrates model performance with a Bland-Altman plot, where the six data points used for model construction are marked with vertical lines and excluded from evaluation. The results show good agreement between the estimation model and the experimental flights, with the model's validation dataset yielding a mean error of 11 W $\pm$ 118 W corresponding to 0.97\% of the full power range (1137.6 W) for the LineDrone, therefore answering the first part of \textbf{Q4}. To validate the second part of \textbf{Q4}, the model was applied to real-world flight trajectories, similar to those shown in Fig. \ref{fig:empirical_power_consumption_vs_model_estimation_for_safety_traj} and Fig. \ref{fig:empirical_power_consumption_vs_model_estimation_for_energy_traj}. The estimations returned by the model closely follow the real power consumption trends observed during flights. For the safest trajectory, the model's absolute mean error was 160.9 W (14\% of the full power range), and for the energy-efficient trajectory, the error was 74.7 W (6.6\% of the same range). In terms of total energy, the safety-oriented trajectory consumed 131.44 kJ in real flights, with a model-estimated error of 4.27\%, whereas the energy-efficient trajectory consumed 91.58 kJ with a model-estimated error of 2.26\%. These results show that the estimation model is sufficiently accurate to estimate power consumption, enabling the optimizer to generate more energy-efficient trajectories. 

\section{Conclusion}
\label{sec:conclusion}


This paper presents ARENA, a risk-aware MOPP framework for UAV inspection missions that combines a genetic optimizer, a voting algorithm to continuously adjust planning behavior according to evolving mission risks, a continuous 4D trajectory representation, and an energy consumption model. Its main contributions are: (a) a voting algorithm for selecting trajectories based on operational risks, (b) an extension of the non-insertion cost proposed in \cite{petit2024}, (c) a 4D continuous trajectory representation enabling speed optimization under actuator constraints, and (d) a semi-empirical energy model accounting for different flight conditions.


Simulation and real-world experiments demonstrate the framework's ability to generate diverse, near-optimal trajectories while balancing safety, energy consumption, and mission duration. It spans the full benchmark ranges for path duration, energy consumption, and average obstacle distance. The proposed energy model estimates instantaneous power with, at worst, a mean error of 14\% and enables trajectories that are up to 30\% more energy-efficient than safer alternatives.

Future work will explore learning-based approaches that continuously estimate energy consumption from velocity vectors, external disturbances, and simplified dynamic models. Such approaches could capture the nonlinear relationship between motion and energy consumption, which analytical models struggle to describe, and thus improve accuracy in real flights. Deep learning techniques will also be investigated to accelerate inference during optimization. Finally, the framework will be generalized to other robotic platforms, such as ground and nautical systems, to demonstrate its versatility across applications.



\bibliographystyle{IEEEtran}
\bibliography{reference}

\end{document}